\documentclass{article}

\usepackage{arxiv}

\usepackage[utf8]{inputenc} 
\usepackage[T1]{fontenc}    
\usepackage{hyperref}       
\usepackage{url}            
\usepackage{booktabs}       
\usepackage{amsfonts}       
\usepackage{nicefrac}       
\usepackage{microtype}      
\usepackage{lipsum}
\usepackage{graphicx}
\usepackage{soul}
\usepackage{bbold}
\soulregister\cite7
\soulregister\ref7
\soulregister\pageref7
\usepackage{algorithm,algorithmic}
\usepackage{amsmath}

\title{Imitation-Based Active Camera Control with Deep Convolutional Neural Network}

\author{
  Christos Kyrkou\thanks{ckyrkou@gmail.com,www.christoskyrkou.com}, \\
  \textit{KIOS Research and Innovation Center of Excellence}\\
  University of Cyprus\\
  1 Panepistimiou Avenue, Nicosia Cyprus \\
  \{kyrkou.christos\}@ucy.ac.cy \\
}

\newcommand{\netname}{ACDCNet}

\newcommand{\netnamespace}{\netname \hspace{1 mm}}

\begin{document}
\maketitle

\begin{abstract}
The increasing need for automated visual monitoring and control for applications such as smart camera surveillance, traffic monitoring, and intelligent environments, necessitates the improvement of methods for visual active monitoring. Traditionally, the active monitoring task has been handled through a pipeline of modules such as detection, filtering, and control. In this paper we frame active visual monitoring as an imitation learning problem to be solved in a supervised manner using deep learning, to go directly from visual information to camera movement in order to provide a satisfactory solution by combining computer vision and control. A deep convolutional neural network is trained end-to-end as the camera controller that learns the entire processing pipeline needed to control a camera to follow multiple targets and also estimate their density from a single image. Experimental results indicate that the proposed solution is robust to varying conditions and is able to achieve better monitoring performance both in terms of number of targets monitored as well as in monitoring time than traditional approaches, while reaching up to 25 FPS. Thus making it a practical and affordable solution for multi-target active monitoring in surveillance and smart-environment applications.
\end{abstract}


\section{Introduction}
Active vision systems (i.e., movable cameras with controllable parameters such as pan and tilt) have received much attention in recent years due to their extended coverage, flexibility, cost-efficiency compared to static vision systems \cite{Micheloni2010VideoAnalysisPTZ}. Active cameras can be used to track targets (i.e. follow them) and can provide continuous monitoring of an area reliably and robustly and are increasingly being used for various applications ranging from surveillance\cite{kyrkou:SCN:TCSVT:2018} to intelligent interactive environments\cite{Wang:VisualComputer:2016}. However, there is a limited number of cameras that human operator can monitor and control hence, there is a need for for automated, robust, and reliable systems for active vision. In addition, there are additional challenges for systems operating with battery limitations or in remote locations and with requirements for rapid deployment in temporary installations necessitating reduced computational cost and simpler active vision systems.


Existing approaches for automated active vision decompose the active control problem into separate modules, such as detection, tracking, and control \cite{Chen:NovelPTControl:2014}, and employ different algorithms at each step. Such examples include motion detection, background modelling/subtraction, and lastly tracking by detection. The former two are widely used in static camera scenarios \cite{Hemangi:2015:2248-9622:138} due to their relative computational efficiency however, for active cameras that move and exhibit constant background change the latter methods are preferred since they are more widely applicable. Such techniques are often augmented with post-processing filters that increase the computational complexity and may also susceptible to false alarms and outliers. Hence, existing methods that rely on hand-crafted features, motion models, and modeling of the camera views are not optimized for active vision scenarios.

\begin{figure}[t]
	\centering
	\includegraphics[width=0.75\linewidth]{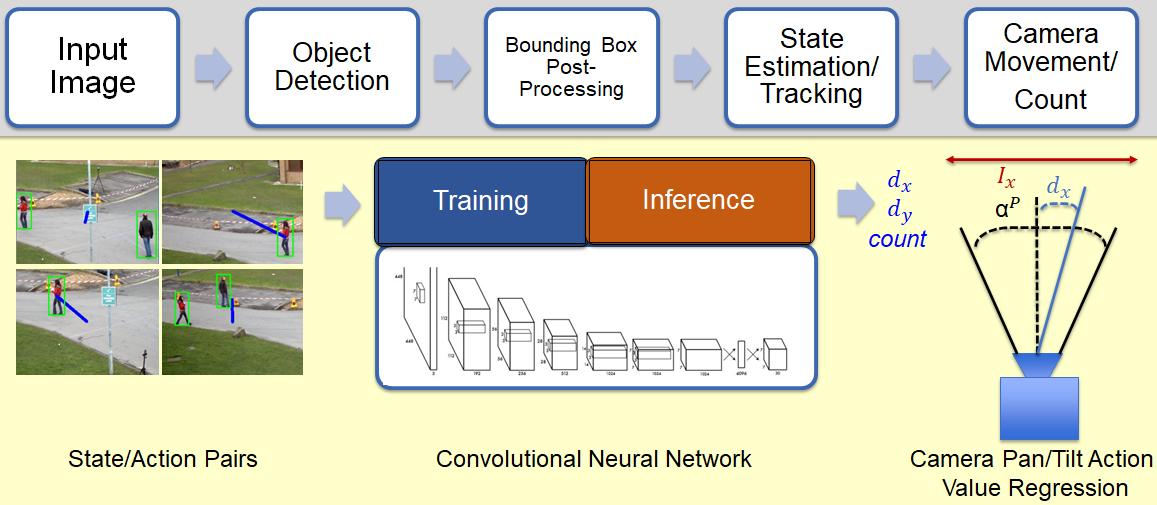}
	\caption{(top) Traditional multi-stage active monitoring vs. (bottom) End-to-end control with deep neural network (DNN). The DNN learns though examples how to control the camera to simultaneously recognize targets and keep them in its Field-of-View (FoV).}
	\label{fig:approach_overview}
\end{figure}

To deal with the aforementioned challenges in this paper we propose leveraging Convolutional Neural Networks (CNN) and end-to-end imitation-learning to develop an active vision system for monitoring people in surveillance applications. To the best of our knowledge there has not yet been any attempt to deal with active tracking of multiple targets in an end-to-end way. In this work, we investigate it and present its appealing potential. In particular an \textbf{A}ctive \textbf{C}amera \textbf{D}eep \textbf{C}ontroller Network, referred to as \netname, is trained in an end-to-end manner through imitation learning to associate single image features with control actions. End-to-end learning approaches allow encapsulating all the intelligence into the machine learning algorithm thus can optimizing all processing steps simultaneously and learning the features to associate with camera control for visual active tracking. In addition, it can result into smaller less complex smart camera systems.

The proposed approach has been verified by extensive experiments using simulation to replicate the motion of the camera and targets. Results indicate that the proposed deep active monitoring approach outperforms some representative methods, composed of several submodules including object detection and tracking, for following multiple targets.

\section{Background and Related Work}

The aim of static object tracking is to localize an object in successive video frames given an annotation the initial frame. It has gained more attention due to its relatively simpler problem setting \cite{Feichtenhofer2017DetTrkDetTrk}. However, the video frames are considered still and hence the approach is not applicable to active cameras where in addition to the challenges associated with moving objects and changing scene perspectives, there is also the issue of controlling the camera and optimize its position for the given vision task. Since the camera itself is constantly changing its orientation, object initialization and continuous tracking pose great challenges. Standard techniques such as foreground segmentation, motion detection or optical flow may face difficulties as the background is constantly changing.

The goal of active monitoring in contrast to a static monitoring setting is to change the camera control parameters in order to maximize a visual-task-related performance objective such as following one or more targets that are located within the field-of-view (FoV), in order to improve the overall surveillance and monitoring capabilities \cite{Bhanu:2011:DVSBook}. Active cameras have gained considerable interest in recent years\cite{Haj:Beyond_Static:2011}, however, there still does not exist a generic solution. Conventional solutions for active visual monitoring tackle the problem by decomposing it into two or more sub-tasks \cite{Ding:Opportunistic:TCSVT:2017}., i.e., object detection typically using a machine-learning-based classifier/detector, a tracking algorithms such as Kalman filter, and a control output for the camera movement. In such case, each task is optimized individually resulting in highly complex systems with many tuning parameters. Furthermore, this leads to difficulty in obtaining real-time performance on resource-constraint embedded camera systems. Different works have investigated the use of active cameras with one or more degrees of freedom to address the problem of monitoring an area and are summarized next. 

Initial approaches such as \cite{Lim2003ImageBasedPT}, followed a master-slave approach used to track targets at a high resolution. One camera, the master, has a wide FoV and performs blob detection and uses a Kalman filter to track a target in an area and by projecting from image plane to world coordinates it controls (pan, tilt, and zoom) the other active camera, the slave, to follow a target. In contrast our goal in this work is to improve the tracking performance of a single camera agent so that it can autonomously follow the majority of targets in its FoV.

In \cite{Haj:reacttivePTZ:2010} an Extended Kalman filter is used to jointly track the object position in the real world as well as estimating the intrinsic camera parameters. The filter outputs are used as inputs to two PID controllers (one for pan and one for tilt motion axis) which continuously track a moving target at a certain resolution. The focus of this work is on tracking only a single target however, and the objects are assumed to have a predetermined size. 

The active camera system proposed in \cite{Wang:VisualComputer:2016} is composed of multiple components in order to track a subject. The face of the target is identified through a face detection system and then a tracker is employed to estimate the targets motion across frames based on previous observations. An online learning approach is also used to learn the appearance model over time and reduce false detections. In addition, Gaussian Mixture Models are used to model the body movement in case the face detection fails. The final part is the controller of the pan-tilt camera which makes its decision of how to move based on where in the image the target is positioned and using a set of rules. Such multi-component system can be difficult to tune and transfer to more resource-constraint systems that monitor multiple targets.

Finally, in \cite{Luo:endtoendRL:2017} an end-to-end active tracker is proposed for following a character in the VizDoom video game environment. It uses reinforcement learning to train an CNN with LSTM to output movement actions for following a single target. Even though they are somewhat realistic, these scenarios do not correspond to real-world use cases. Furthermore, the step-based output is not suitable for the dynamic range needed to control a pan-tilt camera.

In summary, it is evident from the literature that related works make excessive use of multiple modules composed of hand-crafted models and rules that must be tuned separately and in most cases track only a single target. While there has been considerable progression in utilizing deep learning for static camera tracking there has been relatively few works dealing with deep learning for active smart camera systems. Hence, in this work we attempt to bridge the gap between the use of active cameras with deep learning algorithms by proposing an end-to-end learning approach to simultaneously build a detector and controller for cameras with pan and tilt motion capabilities.

\section{Active Camera Control with Deep Learning}
\subsection{Imitation Learning Approach}\label{sec:approachoverview}
To solve the problem of active visual control for target following we frame the problem in an imitation learning setting to be solved in a supervised way. Under this setting the agent (learner) needs to come up with a policy whose resulting state-action behaviour matches the expert behaviour. \footnote{Such expert behaviour can be gathered by appropriate datasets with bounding box annotations for targets from which the camera movement can be extracted.}. For this specific problem the behaviour is how to control the camera given an input frame to keep most of the targets in the field-of-view (FoV). It is assumed that expert state-action pairs are available, ($s_1,a_1$),($s_2,a_2$),\dots,($s_N,a_N$) where $s_i$ are the image frames and $a_i$ are the associated camera actions, such that can form a training set. The expert behaviors are assumed to be i.i.d and then train a regressor network end-to-end in a supervised manner, to learn a control function $f$ that maps observations to actions such that $a_i \simeq f(s_i)$. In doing so, the desired actions are directly regressed during inference, without having to explicitly train the network to first detect the targets (Fig. \ref{fig:approach_overview}).

A pinhole model \cite{Haj:Beyond_Static:2011} is assumed for the camera, and the desired actions are to control its servo motors along its pan and tilt axes (denoted by $a^P$ and $a^T$). Each image $s_i$ has resolution $I_x\times I_y$. Under these circumstances the pixel distances between the image center to the center of mass of the targets is analogous to the angle that the servo motors have to move in order to position the target(s) at the center of FoV (as shown in Fig. \ref{fig:approach_overview}) \cite{Chen:NovelPTControl:2014,kyrkou:SCN:TCSVT:2018}. In addition, the motion of the camera when tracking targets in its FoV is bounded by the half of its horizontal and vertical angles of view (denoted as $\theta_x,\theta_y$). Accordingly, the motion of the camera in the pan and tilt axes can be calculated with respect to the viewing angles. This can be done by first calculating the distance of the camera center from the target(s) center of mass for the horizontal and vertical direction ($d_x,d_y$), and then use Eq. \ref{eq:Motion_Est1} and \ref{eq:Motion_Est2} to associate the pixel distance to the camera angles as also illustrated in (Fig. \ref{fig:approach_overview}). Hence, during the learning process the objective is to map the input image to normalized pixel displacement values ($d_x/I_x,d_y/I_y$), which will then be used to calculate the corresponding servo motor angle displacements in the pan and tilt axis. Predicting offsets instead of angles simplifies the problem and makes it easier for the network to learn. Furthermore, it decouples the learning process from camera specific parameters.

\begin{align}
a^P \rightarrow \dfrac{d_x}{I_x} \times \dfrac{\theta_x}{2}
\label{eq:Motion_Est1}
\end{align}

\begin{align}
a^T \rightarrow \dfrac{d_y}{I_y} \times \dfrac{\theta_y}{2}
\label{eq:Motion_Est2}
\end{align}

\begin{figure}[t]
	\centering
	\includegraphics[width=0.75\linewidth]{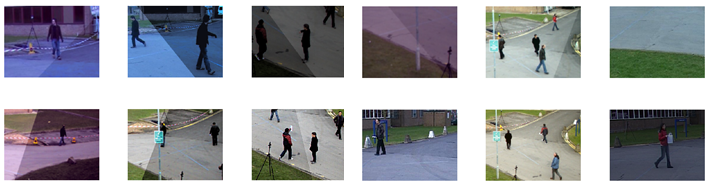}
	\caption{Example of augmented images from the dataset.}
	\label{fig:dataset}
\end{figure}

\subsection{Data Collection and Training}\label{sec:simframe}

Proper training and testing data are necessary to train a deep CNN regressor for the visual active monitoring task. To the best of the author's knowledge there is no publicly available dataset for active vision surveillance applications with ground truth camera controls. For this reason existing wide surveillance image databases, such as the PETS2009\cite{PETS2009}, used for static tracking, are re-purposed to develop a simulation framework that allows for acquiring state-action pairs corresponding to an image frame with the action in image space needed to position the camera to the center-of-mass of the targets in the frame. These ground truth information that can be used for both training and evaluation. In addition, it allows for simulating active cameras and evaluating performance with real-world images, and capturing and storing multiple frame sequences with expert ground truth data for bounding box, density, and camera control. The simulation framework also simulates the movement of a virtual active camera with pan-tilt control with a fixed FoV that is able to move within a larger image frame. 
\begin{figure}[t]
	\centering
	\includegraphics[width=0.75\linewidth]{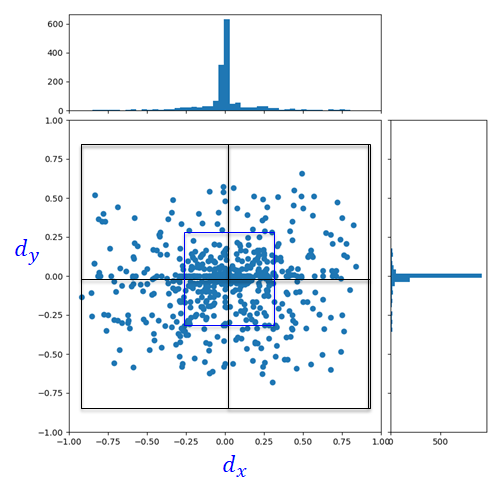}
	\caption{Distribution of the pan and tilt values from the images in the dataset.}
	\label{fig:angledistribution}
\end{figure}

\begin{figure*}[t]
	\centering
	\includegraphics[width=0.75\linewidth]{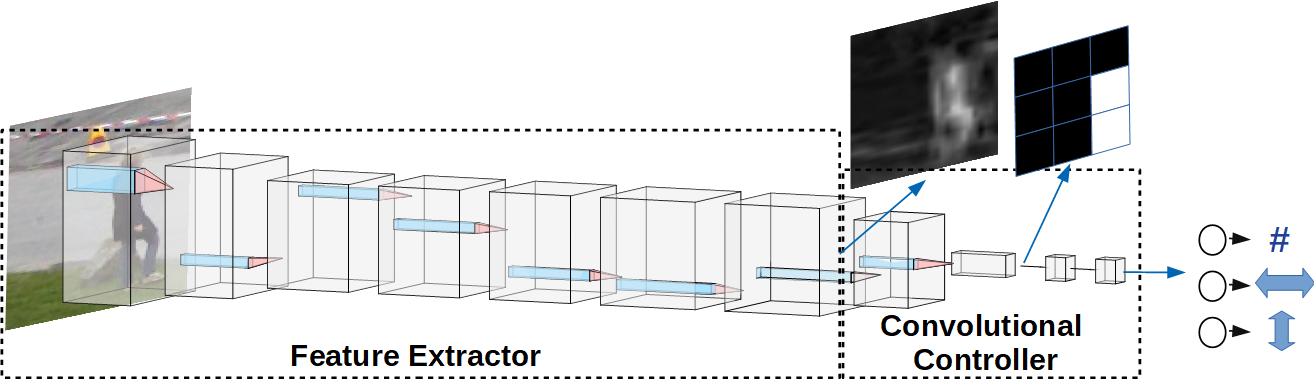}
	\caption{\netnamespace Architecture: Conceptually comprised of a feature extractor and a controller that condenses the high level semantic information to infer an action. Then it regresses the motion for the camera head along the pan and tilt axes as well as the number of targets in the frame.}
	\label{fig:CNN}
\end{figure*}

An initial dataset of over $3100$ images was generated using the simulation software and sequence from the PETS2009\cite{PETS2009} database in order to train \netname. Starting at various positions in random frames in the sequence we calculate the changes in displacement of pan and tilt axes according to Eq. \ref{eq:Motion_Est1} and \ref{eq:Motion_Est2} so that the targets will be positioned at the camera image center in the next frame. Hence, the training set effectively contains single images sampled from the sequence, paired with the corresponding pan and tilt displacement that must be applied. The pan and tilt displacement values are normalized between $-1$ and $1$ with regards to the maximum motion of the camera in both axes. As such, it is also possible to apply any mapping function to scale the output depending on the required configuration. It is also possible to extract different resolution images with varying number of targets in each one, For this work images of $320\times240$ are targeted. It is worth noting that any input image size can be processed after being re-sized appropriately and from cameras with different angles of view. Examples of some of the images in the dataset are shown in Fig. \ref{fig:dataset}

An important part of data preparation is balancing the data. During the data collection process depending on the motion of successive frames, the target displacement values may not change significantly in both axes. Also successive frames may have high correlation and hence little to no motion in both axes. As a result, as shown in Fig. \ref{fig:angledistribution} the distribution of camera motion values might be imbalanced. To ensure that the learning algorithm does not overfit to the majority of values around zero, the data in each batch is sampled to contain high as well as low values so that the network better learns when and how to move. 

The whole data is split into $75\%$ for training the CNN regression model, and $25\%$ of the data was used for testing the final accuracy of the model as well as to compare with other methods. Furthermore augmentations are probabilistically applied to the images during training, to increase the variability and combat over-fitting. The augmentation strategy included some transformations on the image pixels such as blurring and sharpening, color-shifting, illumination changes as well as changes in the image orientation such as translations and horizontal flip with appropriate adjustment of the target pan and tilt values if a geometric augmentation is applied. The combinations of all these augmentations resulted in a variety of novel images that were used for training.

\subsection{Active Camera Deep Controller Network (\netname)}

\netnamespace is trained to perform the regression tasks of estimating the camera motion displacement and the number of targets in the frame. Figure \ref{fig:CNN} shows the network architecture which is conceptually comprised of two main parts, a feature extractor made up of convolutional layers and a motion controller that summarizes the feature maps in order to produce the final output values. Although it is difficult to distinguish the parts of the network that function primarily as feature extractor and which serve as controller since the system is trained end-to-end. The input to the CNN is a $320\times240$ RGB image which is normalized by dividing the pixels by $255$. 

\textbf{Convolutional Feature Extractor:} The layers were designed to perform feature extraction and were chosen empirically through a series of experiments that varied layer configurations. There are $7$ major blocks each comprised of a convolutional layer with \textit{leaky relu} activation with $\alpha=0.3$ and batch-normalization layer. Furthermore, dropout is applied at the middle of the sub-network to combat overfitting with a rate of $0.2$. Overall, the feature extractor is designed to be inherently computationally efficient to support use in embedded smart cameras for local control and decision making. For this reason the first $3$ layers downsample the image to reduce the computational cost and have a small number of filters, and overall the filters do not exceed $128$. After the first $3$ layers the resolution is not reduced further to improve accuracy. The primary goal is to build a small neural network for on-board processing but alternatively transfer learning techniques using other backbone networks can also be applied.

\textbf{Controller Subnetwork:} The controller subnetwork is comprised of both convolutional as well as fully-connected layers. The idea behind this is that the convolutional layers will condense the information from the feature extractor; then the final convolution will convert the feature map into a vector. It is then followed by $2$ fully connected layers with $100$ and $50$ neurons respectively to map the vector to the motion controls. There is a dropout in between the dense layers and all layers have an \textit{elu} activation function for faster convergence. The output of the controller subnetwork is further processed through a clipped linear activation that bounds the output between $[-1,\dots,1]$ to estimate the motion in the horizontal and vertical direction more effectively. The third output neuron that regresses the number of targets uses a ReLU function to discard negative numbers. Overall, \netnamespace has a total of $\sim386,000$ parameters. This results in a small network which requires $\sim4$MB, resulting in a lightweight network that can run even on low-end CPUs. 

\subsection{Network Training}
The objective of the learning process is to regress two motion values in the pan and tilt axes, and the number of targets. Accordingly, the loss function in Eq. \ref{eq:loss} is employed for learning the camera controls and target count. The \textit{Keras} deep learning framework \cite{keras} with \textit{Tensorflow} \cite{tensorflow} running as the backend is used for the training of the CNN regressor. The network was trained using a GeForce Titan Xp, on a PC with an Intel $i7-7700K$ processor, and $32$GB of RAM. The Adam optimization method was used for training with a learning rate step-decay approach starting from an initial learning rate of $0.001$, and decreasing it by a factor of $0.95$ every $5$ epochs. The CNN regressor is trained for $500$ epochs with a batch size of $32$ resulting in $3100$ generated training images per epoch.

\small{
\begin{align}
 \nonumber L=\dfrac{1}{N_B}\sum^{N_B}_{j=1}\bigg[(y^{true}_{c}(j)-y^{pred}_{c}(j))^2+\\
  \nonumber|y^{true}_{d_{x}}(j)-y^{pred}_{d_{x}}(j)| +  && \\  |y^{true}_{d_{y}}(j)-y^{pred}_{d_{y}}(j)|\bigg], &&
\label{eq:loss}
\end{align}
}

\section{Evaluation and Experimental Results}
To compare the proposed method a traditional active tracking pipeline is implemented following other relevant approaches for active camera control \cite{Ding:Opportunistic:TCSVT:2017}. In this paradigm an object detector is used to localize the targets and a tracker is applied to filter the detections over time. Finally based on the filter target positions the camera is moved to position the targets as close to the center as possible. The control is performed using the simulation framework as discussed in Section \ref{sec:simframe} and is the same for all methods.  

We use the YOLO \cite{YOLOv2} framework to  comparison with single-shot detectors which are considered more suitable for real-time applications and can be leveraged for real-time embedded systems. We show herein comparisons with the orignal smaller YOLO variant referred to as \textit{tinyYOLO}. Comparisons are also made against the widely used SVM-HOG pedestrian detection algorithm \cite{Dalal:Triggs:HOG:2005} implemented in OpenCV \cite{opencv_library}  and still favored for embedded applications and  especially suitable for low-power systems \cite{kyrkou:SCN:TCSVT:2018}. On top of the detectors we apply tracking with Kalman filter (denoted with \textit{\_T} in the Figures) which is a common approach used in active tracking \cite{Haj:Beyond_Static:2011,Haj:reacttivePTZ:2010}. The incorporated filtering handles bounding box associations, maintains trajectories, and handles the creation and termination of tracks.

\begin{figure}[t]
	\centering
	\includegraphics[width=0.75\linewidth]{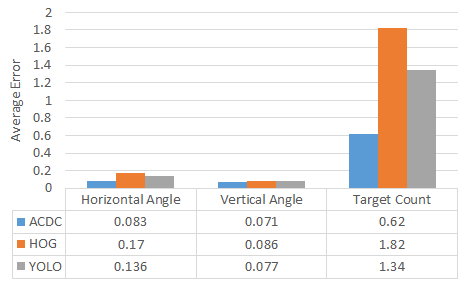}
	\caption{Errors for the different regressed quantities for each approach.}
	\label{fig:errors}
\end{figure}

The proposed model is evaluated experimentally in a number of ways to determine its overall performance. For the first experiment we use a test set of $500$ images not used in the training process, with ground truth camera motion information in order to calculate how accurate is each method in estimating the camera motion controls for positioning targets at the center of its FoV as well as regressing the number of targets. As shown in Fig. \ref{fig:errors} the proposed CNN achieves the lowest error across the different methods. In practice this means that it will follow the targets better since it has not have to deal with bounding box localization errors that can steer off the camera control. In particular, for the case of estimating the number of targets the other methods can be as off as 1-2 targets. Furthermore, the lower errors also mean that in cases where there is no target in its FoV the network has recognized that it should not move, and in case of many targets it attempts to follow their center of mass thus keeping the majority of them in view.

The next experiment involved evaluating all the aforementioned methods in the simulation environment and measure important metrics such as the average number of targets followed. This provides an indication of how well each approach manages to keep up with the target motion as well as how many targets it can keep in its field of view. To conduct these sequences where generated from the PETS2009\cite{PETS2009} dataset for which the ground truth bounding box annotations is available but have not been used in the training and validation phases. These sequences provide an additional challenge as they feature different visual conditions with high density crowd densities and motion patterns. The images in the sequences are of $768\times576$ resolution and the virtual camera FoV is set to $320\times240$ for all methods so that there is margin for the camera to move and follow the targets.

The developed simulation framework outlined in section \ref{sec:simframe} models the camera motion based on the visual input and evaluates its performance. The output of each vision pipeline is a motion vector that will be passed to the simulator to perform the action. The camera FoV for all methods is set at the same initial position. In all cases the objective of the camera is to move in such a way as to keep the most number of targets in its FoV. We also report the ground truth motion which represents the motion of the camera when the target positions are perfectly known.

\begin{figure}[t]
	\centering
	\includegraphics[width=0.75\linewidth]{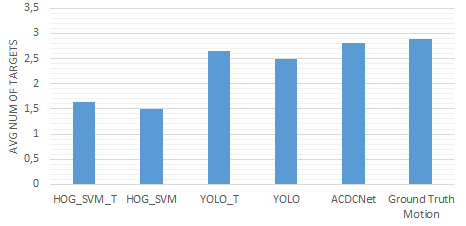}
	\caption{Average number of targets monitored over time for each approach.}
	\label{fig:number_tars}
\end{figure}

\begin{figure}[t]
	\centering
	\includegraphics[width=0.75\linewidth]{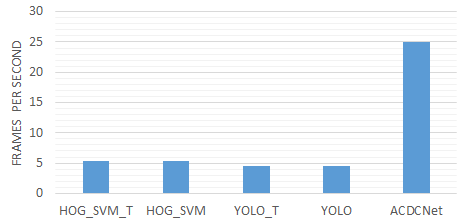}
	\caption{Frames-per-Second (FPS) achieved by each approach.}
	\label{fig:FPS}
\end{figure}

\begin{figure}[t]
	\centering
	\includegraphics[width=0.75\linewidth]{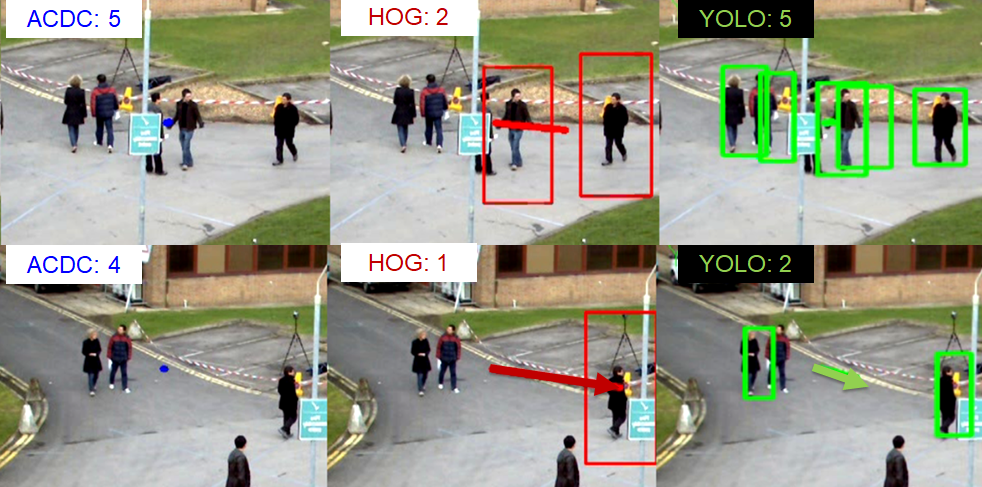}
	\caption{Example of detection/predictions and estimation for number of targets. A common mistake by HoG-SVM and YOLO involves missing targets. In addition, other source of mistake relate to wrong localization with the bounding box. In both cases leads the camera to move in the wrong direction. Finally \netnamespace is able to correctly regress the number of targets and chooses to not move the camera in these frames as the targets are mostly in its FoV. (Best viewed in color)}
	\label{fig:img_errors}
\end{figure}

\begin{figure}[t]
	\centering
	\includegraphics[width=0.75\linewidth]{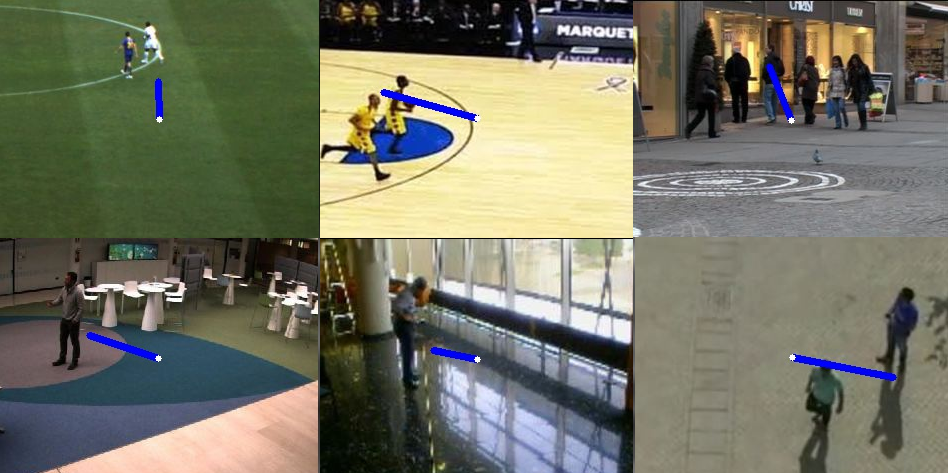}
	\caption{Example of applying the approach to real-world images. The motion vector shows the predicted motion that the camera will make to keep the targets in its field of view. (Best viewed in color)}
	\label{fig:real_world}
\end{figure}

The results are presented in Fig. \ref{fig:number_tars} and Fig. \ref{fig:FPS}. The proposed system manages to surpass the other approaches and be close to the optimum based on the number of targets tracked over the length of the sequences. The proposed approach also manages to have a performance of $\sim$ 25 frames per second on a CPU. Since this is a simulation the effect of processing time was not evaluated directly. However, since the inference time of the other methods is higher it is expected that in real conditions slower methods will miss even more targets. Hence, the proposed approach provides a much more efficient way of building reactive active vision systems that can also be more responsive to the target motion. Overall the performance gains between the proposed and other methods can be attributed end-to-end nature of our approach that associates lower-level-features with control actions and does rely on bounding boxes. As a result \netnamespace manages to handle more dense targets better. Fig. \ref{fig:img_errors} demonstrates some cases that cause errors in the traditional detection approaches; while Fig. \ref{fig:real_world} also demonstrates the fact that the approach can generalize in out of domain images gathered from real-world settings.

\section{Conclusion}
This work addresses the problem of controlling a smart camera for active vision applications (surveillance, drones, vehicles, smart environments). One of the major contributions is that the problem is tackled via end-to-end learning using deep convolutional neural networks. A CNN architecture referred to as \netnamespace is proposed that maps input images to pan and tilt motion command with the goal of keeping targets in its field-of-view. Even with single frame information the proposed network outperforms other multi-stage approaches in terms of monitoring efficiency and provides higher frame-rates while being lightweight. The results are promising and as future work we will further explore what the network actually learns through its internal representations, and deploy it in a real experimental setup for further evaluation.

\section*{Acknowledgment}
This work was supported by the European Union's Horizon 2020 research and innovation programme under grant agreement No 739551 (KIOS CoE) and from the Government of the Republic of Cyprus through the Directorate
General for European Programmes, Coordination and Development. 

The author would like to acknowledge the support of NVIDIA Corporation with the donation of the Titan Xp GPU used for this research.

\bibliographystyle{IEEEtran}
\bibliography{IEEEabrv,references}

\end{document}